\documentclass{article}
\usepackage{microtype}
\usepackage{graphicx}
\usepackage{subfigure}
\usepackage{booktabs} 
\usepackage{hyperref}

\usepackage[accepted]{icml2018}
\usepackage{latexsym}
\usepackage{amsfonts,amssymb}
\usepackage{amsmath}
\usepackage{graphicx}
\usepackage{algorithm}
\usepackage{algorithmic}
\usepackage{float}
\usepackage{array}
\usepackage{multirow}
\usepackage{url}
\usepackage{makecell}

\begin{document}

\twocolumn[
\icmltitle{NDT: Neual Decision Tree Towards Fully Functioned Neural Graph}

\begin{icmlauthorlist}
\icmlauthor{Han Xiao}{thu}
\end{icmlauthorlist}

\icmlaffiliation{thu}{State Key Laboratory of Intelligent Technology and Systems, National Laboratory for Information Science and Technology, Department of Computer Science and Technology, Tsinghua  University, Beijing 100084, PR China}

\icmlcorrespondingauthor{Han Xiao}{Almighty.Xiao.Han@iCloud.com}

\icmlkeywords{logics Empowered, Neural Architecture}

\vskip 0.3in]

\printAffiliationsAndNotice{}  

\begin{abstract}
Though traditional algorithms could be embedded into neural architectures with the proposed principle of \cite{xiao2017hungarian}, the variables that only occur in the condition of branch could not be updated as a special case. To tackle this issue, we multiply the conditioned branches with Dirac symbol (i.e. $\mathbf{1}_{x>0}$), then approximate Dirac symbol with the continuous functions (e.g. $1 - e^{-\alpha|x|}$). In this way, the gradients of condition-specific variables could be worked out in the back-propagation process, approximately, making a fully functioned neural graph. Within our novel principle, we propose the neural decision tree \textbf{(NDT)}, which takes simplified neural networks as decision function in each branch and employs complex neural networks to generate the output in each leaf. Extensive experiments verify our theoretical analysis and demonstrate the effectiveness of our model.
\end{abstract}

\section{Introduction}
Inspired by brain science, neural architectures have been proposed in 1943, \cite{Mcculloch1943A}. This branch of artificial intelligence develops from single perception \cite{Casper1969Perceptrons} to deep complex network \cite{Lecun2015Deep}, achieving several critical successes such as AlphaGo \cite{Silver2016Mastering}.  Notably, all the operators (i.e. matrix multiply, non-linear function, convolution, etc.) in traditional neural networks are numerical and continuous, which could benefit from back-propagation algorithm, \cite{Rumelhart1988Learning}. 

Recently, logics-based methods (e.g. Hungarian algorithm, max-flow algorithm, A$^*$ searching) are embedded into neural architectures in a dynamically graph-constructing manner, opening a new chapter for intelligence system, \cite{xiao2017hungarian}. \textit{\textbf{Generally, neural graph is defined as the intelligence architecture, which is characterized by both logics and neurons.}}

With this proposed principle from the seminal work, we attempt to tackle image classification. Specifically, regarding this task, the overfull categories make too much burden for classifiers, which is a normal issue for large-scale datasets such as ImageNet \cite{deng2009imagenet}. \textit{\textbf{We conjecture that it would make effects to roughly classify the samples with decision tree, then category the corresponding samples with strong neural network in each leaf, because in each leaf, there are much fewer categories to predict.}} The attribute split in traditional decision trees (e.g. \textit{ID3, Random Forest, etc.}) is oversimplified for precise pre-classification, \cite{Zhou2017Deep}. Thus, we propose the method of neural decision tree \textbf{(NDT)}, which applies neural network as decision function to strengthen the performance.

\begin{figure}
	\centering
	\includegraphics[width=0.9\linewidth]{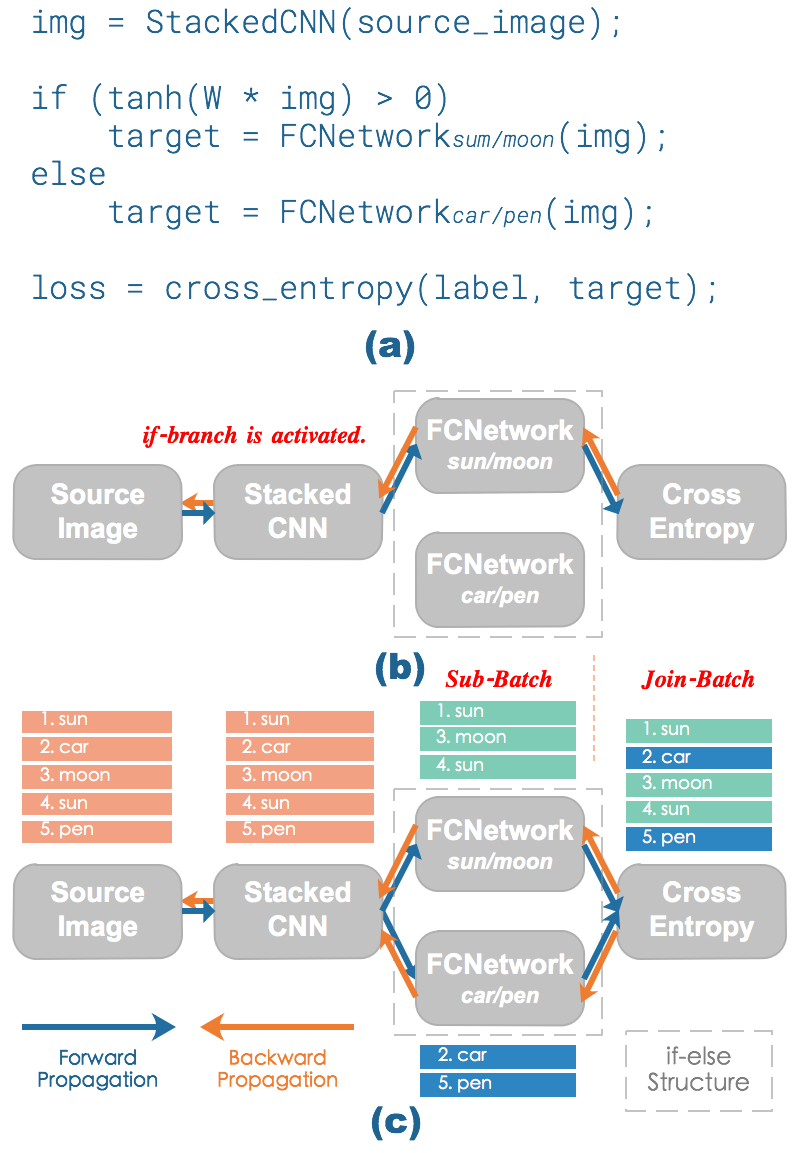}
	\caption{Illustration of how logic flow is processed in our methodology. Referring to (b), we process the \textit{if-else} structure of (a) in a dynamically graph-constructing manner. Theoretically, we construct the graph according to the active branch, in the forward propagation. When the forward propagation has constructed the graph according to logic instructions, the backward propagation would be performed as usual in a continuous and non-branching graph. Practically, the dynamically constructed process corresponds to batch operations. The samples with id \textit{1,3,4} are activated in \textit{if} branch, while those with id \textit{2,5} are tackled in \textit{else} branch. After the end of \textit{if-else} instruction, the sub-batch hidden representations are joined as the classified results.}
	\label{fig:basicidea}
\end{figure}

Regarding the calculus procedure of NDT, the basic principle is to treat the logic flow (i.e. \textit{``if, for, while'' in the sense of programming language}) as a dynamic graph-constructing process, which is illustrated in Figure \ref{fig:basicidea}.

This figure demonstrates the classification of four categories (i.e. \textit{sun, moon, car and pen}), where an \textit{if} structure is employed to split the samples into two branches (i.e. \textit{sun-moon, car-pen}), where the fully connected networks generate the results respectively. In the forward propagation, our methodology activates some branch according to the condition of \textit{if}, then dynamically constructs the graph according to the instructions in the activated branch. In this way, the calculus graph is constructed as a non-branching and continuous structure, where backward propagation could be performed conventionally, demonstrated in Figure \ref{fig:basicidea} (b). Generally, we should note that the repeat (i.e. \textit{for, while}) could be treated as performing \textit{if} in multiple times, which could also be tackled by the proposed principle. Thus, all the traditional algorithms could be embedded into neural architectures. For more details, please refer to \cite{xiao2017hungarian}.

However, as a special challenge of this paper, the variables that are only introduced in the condition of branch could not be updated in the backward propagation, because they are outside the dynamically constructed graph, for the example of $W$ in Figure \ref{fig:basicidea}. Thus, to make a completely functioned neural graph, this paper attempts to tackle this issue in an approximated manner. Simply, we multiply the symbols inside the branch with Dirac function (i.e. $\mathbf{1}_{x>0}$ or $\mathbf{1}_{x\le0}$). Specifically, regarding Figure \ref{fig:basicidea}, we reform $FCNetwork(img)$ as $FCNetwork(img \otimes \mathbf{1}_{tanh>0})$ in the \textit{if} branch and perform the corresponding transformation in the \textit{else} branch, where $\otimes$ is element-wise multiplication. The forward propagation would not be modified by the reformulation, while as to the backward process, we approximate the Dirac symbol with a continuous function to work out the gradients of condition, which solves this issue. It is noted that in this paper, the continuous function is $1 - e^{-\alpha|x|} \approx \mathbf{1}_{x>0}$.

We conduct our experiments on public benchmark datasets: MNIST and CIFAR. Experimental results illustrate that our model outperforms other baselines extensively and significantly, which illustrates the effectiveness of our methodology. The most important conclusion is that ``our model is differentiable'', which verifies our theory and provides the novel methodology of fully functioned neural graph.

\textbf{Contributions} \textbf{(1.)} We complete the principle of neural graph, which characterizes the intelligence systems with both logics and neurons. Also, we provide the proof that neural graph is Turing complete, which makes a learnable Turing machine for the theory of computation. \textbf{(2.)} To tackle the issue of overfull categories, we propose the method of neural decision tree \textbf{(NDT)}, which takes simplified neural networks as decision function in each branch and employs complex neural networks to generate the output in each leaf. \textbf{(3.)} Our model outperforms other baselines extensively, verifying the effectiveness of our theory and method.

\textbf{Organization.}  In the Section 2, our methodology and neural architecture are discussed. In the Section 3, we specific the implementation of fully functioned neural graph in detail. In the Section 4, we provide the proof that neural graph is Turing complete. In the Section 5, we conduct the experiments for performance and verification. In the Section 6, we briefly introduce the related work. In Section 7, we list the potential future work from a developing perspective. Finally in the Section 8, we conclude our paper and publish our codes.

\begin{figure}[t]
	\centering
	\includegraphics[width=0.8\linewidth]{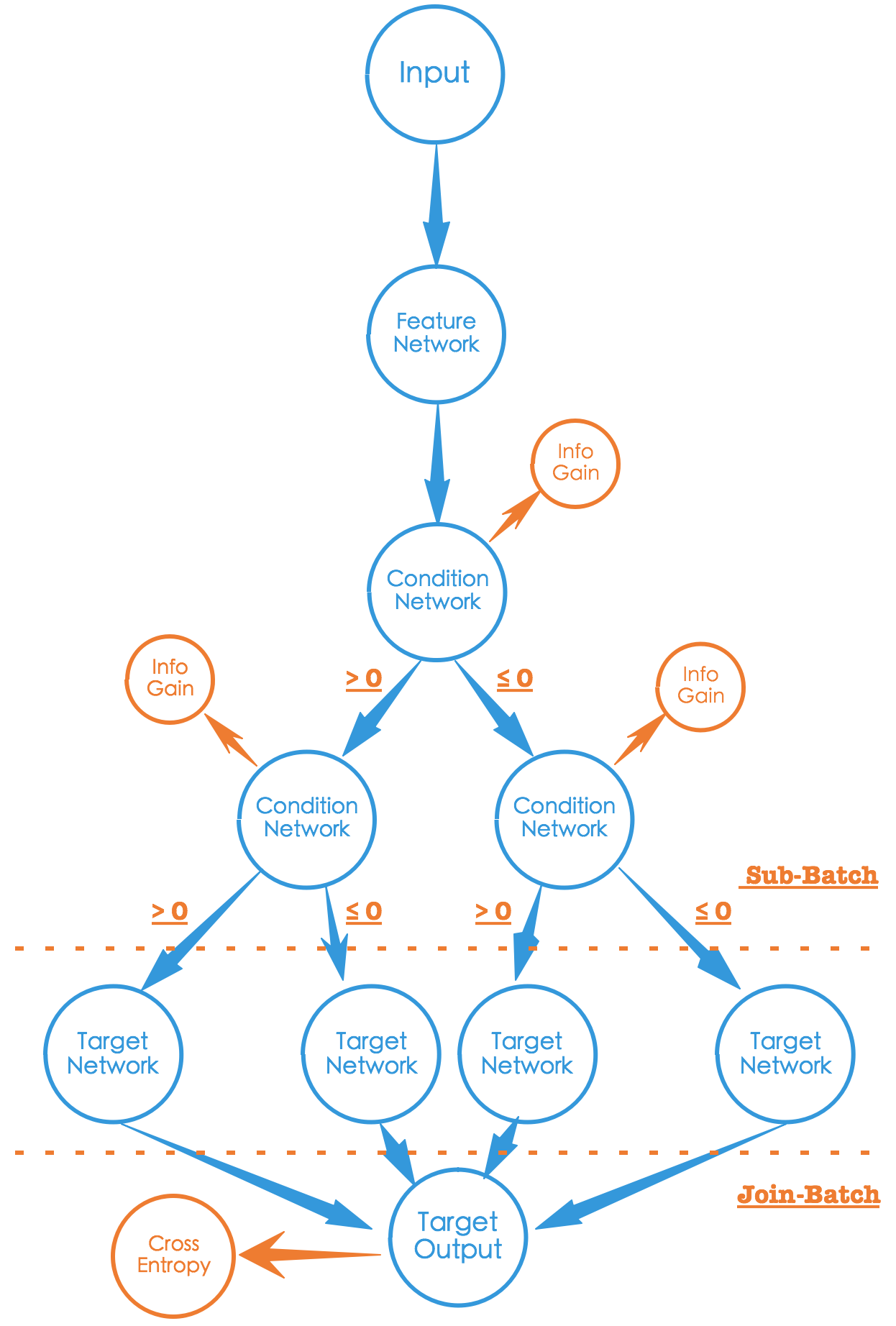}
	\caption{The neural architecture of NDT (depth = 2). The input is classified by decision tree component with the condition networks, then the target networks predict the categories for each sample in each leaf. Notably, the tree component takes advantages of sub-batch technique, while the targets are joined in batch to compute the cross entropy objective.}
	\label{fig:methodology}
\end{figure}

\section{Methodology}
First, we introduce the overview of our model. Then, we discuss each component, specifically. Last, we discuss our model from the ensemble perspective.

\subsection{Architecture}

Our architecture is illustrated in Figure \ref{fig:methodology}, which is composed by three customized components namely feature, condition and target network. Firstly, The input is transformed by feature network and then the hidden features are classified by decision tree component composed by hierarchal condition networks. Secondly, the target networks predict the categories for each sample in each leaf. Finally, the targets are joined to work out the cross entropy objective. The process is exemplified in Algorithm \ref{alg}. 

\textbf{Feature Network.} To extract the abstract features with deep neural structures, we introduce the feature network, which is often a stacked CNN and LSTM.

\textbf{Condition Network.} To exactly pre-classify each sample, we employ a simplified neural network as condition network, which is usually a one- or two-layer multi-perceptions with the non-linear function of $tanh$. This layer is only applied in the inner nodes of decision tree. Actually, the effectiveness of traditional decision tree stems from the information gain splitting rules, which could not be learned by condition networks, directly. Thus, we involve an objective item for each decision node to maximize the information gain as:
\begin{eqnarray}
	\max InfoGain & = & \frac{N_{left}}{N_{total}} \sum_{j=0}^{|F|} p_j^{left} ln(p_j^{left}) + \nonumber \\ 
	& & \frac{N_{right}}{N_{total}} \sum_{j=0}^{|F|} p_j^{right} ln(p_j^{right})
\end{eqnarray} 
where $N$ is the corresponding count, $|F|$ is the feature number and $p$ is the corresponding probabilistic distribution of features. Regarding the derivatives relative to Dirac symbol, we firstly reformulate the information gain in the form of Dirac symbol as:
\begin{eqnarray}
 	N_{left} = \sum_{i=1}^{N_{total}} \mathbf{1}_{cn > 0} \\
 	N_{right} = \sum_{i=1}^{N_{total}} \mathbf{1}_{cn \le 0}
\end{eqnarray}
\begin{eqnarray}
 	p_j^{left} = \frac{\sum_{i=0}^{N_{total}} \mathbf{1}_{cn > 0, j} L_{i,j}}{\sum_{i=0}^{N_{total}} \mathbf{1}_{cn > 0}} \\
 	p_j^{right} = \frac{\sum_{i=0}^{N_{total}} \mathbf{1}_{cn \le 0, j} L_{i,j}}{\sum_{i=0}^{N_{total}} \mathbf{1}_{cn \le 0}} 
\end{eqnarray}
where $cn$ is short for condition network and $L_{i,j}$ is the ad-hoc label vector of $i$-th sample, where the true label position is 1 and otherwise 0. By simple computations, we have:
\begin{eqnarray}
	\frac{\partial IG}{\partial \mathbf{1}_{cn > 0}} = \frac{\sum_{i=0}^{N_{total}} L_{i,j} ln(p_j^{left})}{N_{total}} \\
	\frac{\partial IG}{\partial \mathbf{1}_{cn \le 0}}	= \frac{\sum_{i=0}^{N_{total}} L_{i,j} ln(p_j^{right})}{N_{total}} 
\end{eqnarray}
where $IG$ is short for $InfoGain$. As discussed in Introduction, we approximate the Dirac symbol as a continuous function, specifically as $1 - e^{-\alpha|x|} \approx \mathbf{1}_{x>0}$. Thus, the gradient of condition network could be deducted as:
\begin{eqnarray}
\frac{\partial IG}{\partial cn} \approx \frac{\partial IG}{\partial \mathbf{1}_{cn > 0}} \frac{\partial (1 - e^{-\alpha|cn|)}}{\partial cn} =  \frac{\partial IG}{\partial \mathbf{1}_{cn > 0}} (\alpha e^{-\alpha|x|} s(cn)) \\
\frac{\partial IG}{\partial cn} \approx \frac{\partial IG}{\partial \mathbf{1}_{cn \le 0}} \frac{\partial (1- e^{-\alpha|cn|)}}{\partial cn} =  \frac{\partial IG}{\partial \mathbf{1}_{cn \le 0}} (\alpha e^{-\alpha|x|} s(cn))
\end{eqnarray}
where $s$ is the sign function. 

Actually, all the reduction could be performed automatically within the proposed principle, that to multiply the symbols inside the branch with Dirac function. Specifically, as an example of the count, $N_{left} = \sum_{i=1}^{N_{total}} 1 \times \mathbf{1}_{cn > 0}$.

\textbf{Target Network.} To finally predict the category of each sample, we apply a complex network as the target network, which often is a stacked convolution one for image or an LSTM for sentence. 

\begin{algorithm}
\renewcommand{\algorithmicrequire}{\textbf{Input:}}
\renewcommand{\algorithmicensure}{\textbf{Output:}}
\caption{Neural Decision Tree \textbf{(NDT)}}
\label{alg}
\includegraphics[width=\linewidth]{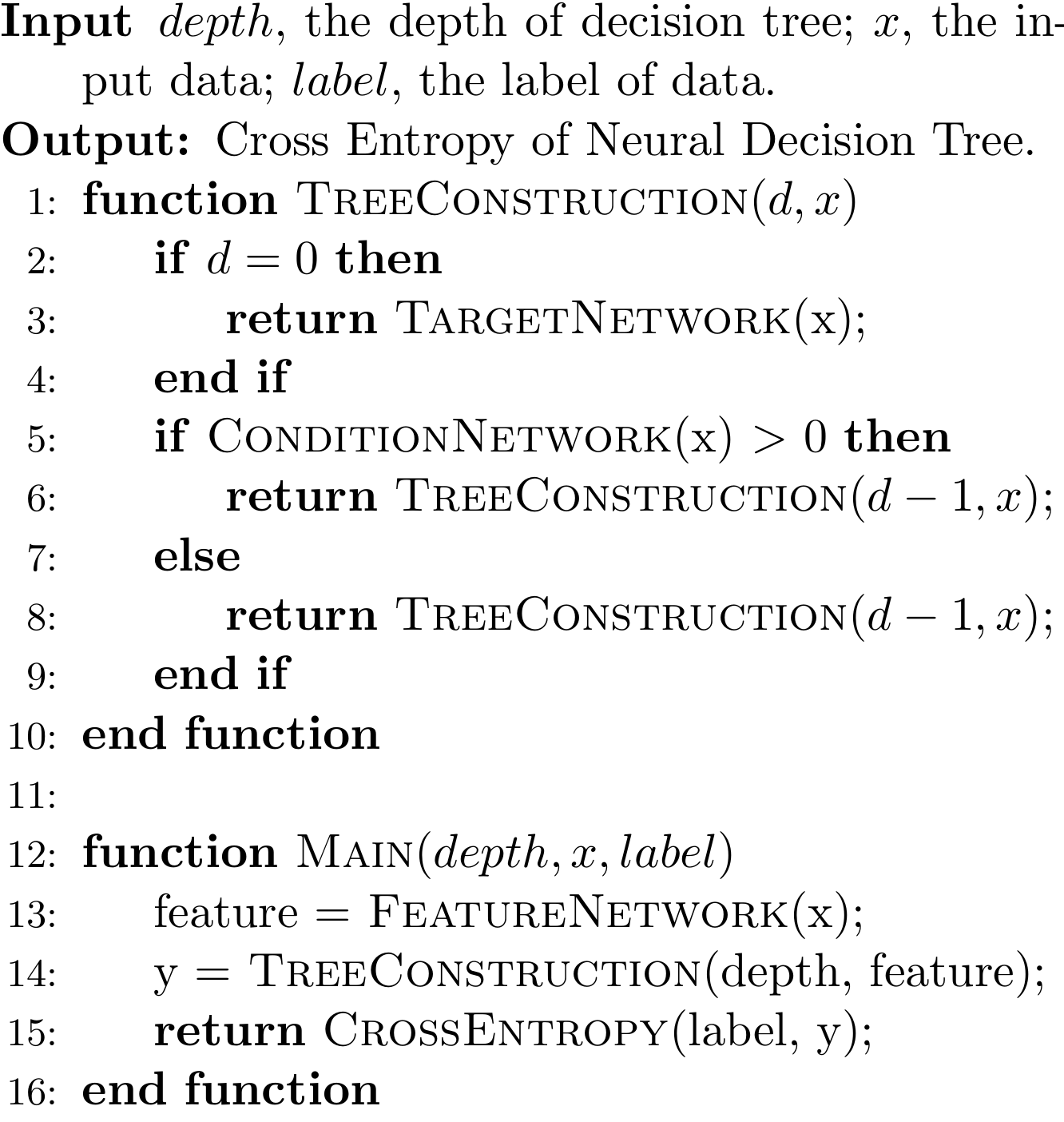}
\end{algorithm}

\subsection{Analysis From Ensemble Perspective}
NDT could be treated as an ensemble model, which ensembles many target networks with the hard branching condition networks. Currently, there exist two branches of ensemble methods, namely split by features, or split by samples, both of which increases the difficulty of single classifier. However, NDT splits the data by categories, which means single classifier deals with a simpler task. 

The key point is the split purity of condition networks, because the branching reduces the sample numbers for each leaf. Relatively to single classifier, if our model keeps the sample number per category, NDT could make more effects. For an example of one leaf, the sample number reduces to 30\%, while the category number reduces from 10 to 3. With similarly sufficient samples, our model deals with 3-classification, which is much easier than 10-classification. Thus, our model benefits from the strengthen of single classifier.

\section{Dynamical Graph Construction} 
Previously introduced, neural graph is the intelligence architecture, which is characterized by both logics and neurons. Mathematically, the component of neurons are continuous functions, such as matrix multiply, hyperbolic tangent (tanh), convolution layer, etc, which could be implemented as mathematical operations. 

Obviously, simple principal implementation for non-batch mode is easy and direct. But practically, all the latest training methods take the advantages of batched mode. Hence, we focus on the batched implementation of neural graph in this section.

Conventionally, neural graph is composed by two styles of variable, namely symbols such as $W$  in Figure \ref{fig:basicidea}, and atomic types such as the integer $d$ in Algorithm \ref{alg} Line 2. In essence, symbolic variables originate from the weights between neurons, while the atomic types are introduced by the embedded traditional algorithms.

Therefore, regarding the component of logics, there exist two styles: symbol- and atomic-type-specific logic components, which are differentiated in implementation. Symbol-specific logics indicates the condition involves the symbols, such as Line 5 $\sim$ 9 in Algorithm \ref{alg}, while atomic-type-specific logics means there are only atomic types in the condition such as Line 2 in Algorithm \ref{alg}. However, our proposed principle, that dynamically constructing neural graph, could process both the situations.

To implement symbol-specific logic component, we propose two batch operations, namely sub- and join-batch operation. Take the example of Figure \ref{fig:basicidea} (c). To begin, there are five samples in the batch. In the forward pass, once processing in the branch, according to the condition, the batch is split into two sub-batches, each of which is respectively tackled by the instructions in the corresponding branch, simultaneously. After processed by two branches, the sub-batches are joined into one batch, according to the original order. In the backward propagation, the gradients of joined batch are split into two parts, which correspond to two sub-batches. When the process has propagated through two branches, the gradients of two sub-batches are joined again to form the gradients of stacked CNN. 

Theoretically, a sample in some sub-batch means the corresponding branch is activated for this sample and the other branch is deactivated. On the other word, the hidden representations of this sample connect to the activated branch rather than the deactivated one. Thus, the symbol-specific logic components perform our proposed principle, in the manner of sub- and join-batch operation. Notably, if there is no variable that is only introduced in the condition, it is unnecessary to update the condition, which makes corresponding neural graph an exact method.

To implement atomic-type-specific logic component, we propose a more flexible batch operation namely allocate-batch. Take the example of Hungarian Layer \cite{xiao2017hungarian}. The Hungarian algorithm deals with the similarity matrix to provide the alignment information, according to which, the dynamic links between symbols are dynamically allocated, shown in Figure 4 of \cite{xiao2017hungarian}. Thus, the forward and backward propagation could be performed in a continuous calculus graph. Simply, in the forward pass, we record the allocated dynamic links of each sample in the batch, while in the backward pass, we propagate the gradients along these dynamic links. Obviously, the atomic-type-specific logic components perform our proposed principle, in the manner of allocate-batch operation.

The traditional algorithms are a combination of branch (i.e. \textit{if}) and repeat (i.e. \textit{for, while}). Repeat could be treated as performing branch in multiple times. Thus, the three batch operations, namely sub-, join- and allocate-batch operation, could process all the traditional algorithms, such as resolution method, A$^*$ searching, Q-learning, label propagation, PCA, K-Means, Multi-Armed Bandit (MAB), AdaBoost.

\section{Neural Graph is Turing Complete}
Actually, if neural graph could simulate the Turing machine, it is Turing complete. Turing machine is composed by four parts: a cell-divided tape, reading/writing head, state register and a finite table of instructions. Correspondingly, symbols are based on tensor arrays, which simulate the cell-divided tape. Forward/Backward process indicate where to read/write. Atomic-type-specific variables record the state. Last, the logic flow (i.e. \textit{if, while, for}) constructs the finite instruction table. \textit{\textbf{In summary, neural graph is Turing complete.}}

Specifically, neural graph is a learnable Turing machine rather than a static one. Learnable Turing machine could adjust the behaviors/performance, according to data and environment. Traditional computation models focus on static algorithms, while neural graph takes advantages of data and perception to strengthen the rationality of behaviors. 

\section{Experiment}
In the section, we verify our model on two datasets: MNIST \cite{Lecun1998Gradient} and CIFAR \cite{Krizhevsky2009Learning}. We first introduce the experimental settings in Section 4.1. Then, in Section 4.2, we conduct performance experiments to testify our model. Last, in Section 4.3, to further verify our theoretical analysis, that NDT could reduce the category number of leaf nodes, we perform a case study to justify our assumption.

\subsection{Experimental Setting}
There exist three customized networks in our model, that the feature, condition and target network. We simply apply identify mapping as feature network. Regarding the condition network, we apply a two-layer fully connected perceptions, with the hyper-parameter input-300-1 for MNIST and input-3000-1 for CIFAR. Regarding the target network, we also employ a three-layer fully connected perceptions, with the hyper-parameter input-300-100-10 for MNIST, input-3000-1000-10 for CIFAR-10 and input-3000-1000-100 for CIFAR-100. \footnote{We know the feature and target network are too oversimplified for this task. But this version targets at an exemplified model, which still could verify our conclusions. We will perform a complex feature and target network in the next/final version.} To train the model, we leverage AdaDelta \cite{Zeiler2012ADADELTA} as our optimizer, with hyper-parameter as moment factor $\eta=0.6$ and $\epsilon=1 \times 10^{-6}$.  We train the model until convergence, but at most 1,000 rounds. Regarding the batch size, we always choose the largest one to fully utilize the computing devices. Notably, the hyper-parameters of approximated continuous function is $\alpha=1000$.

\subsection{Performance Verification}
\textbf{MNIST.} The MNIST dataset \cite{Lecun1998Gradient} is a classic benchmark dataset, which consists of handwritten digit images, 28 x 28 pixels in size, organized into 10 classes (0 to 9) with 60,000 training and 10,000 test samples.  We select some representative and competitive baselines: modern CNN-based architecture LeNet-5 with dropout and ReLUs, classic linear classifier SVM with RBF kernel, Deep Belief Nets and a standard Random Forest with 2,000 trees. We could observe that: 
\begin{enumerate}
	\item NDT will beat all the baselines, verifying our theory and justifying the effectiveness of our model.
	\item Compared to single target network, NDT promotes 0.65 point, which illustrates the ensemble of target network is effective.
	\item Compared to Random Forest that is also a tree-based method, NDT promotes 0.75 point, which demonstrates the neurons indeed strengthen the decision trees.
\end{enumerate}

\begin{table}[t]
	\caption{Performance Evaluation on MNIST Dataset.}
	\centering
	\label{tab:paraphrase}
	\renewcommand\arraystretch{1.1}
	\begin{tabular}{c c}
		\Xhline {1.2pt} Methods & Accuracy (\%)\\
		\Xhline {1.2pt}
		Single Target Network & 96.95 \\
		LeNet-5 & 99.50 \\
		Multi-Perspective CNN & 81.38 \\
		Deep Belief Net & 98.75 \\
		SVM (RBF kernel) & 98.60 \\
		Random Forest & 96.80 \\
		\hline
		\hline 
		\textbf{NDT (depth = 2)} & \textbf{97.90} \\
		\Xhline {1.2pt}
	\end{tabular}
\end{table}

\textbf{CIFAR.} The CIFAR-10/100 dataset \cite{Krizhevsky2009Learning}, is also a classic benchmark for overfull category classification, which consists of color natural images, 32 x 32 pixels in size, from 10/100 classes with 50,000 training and 10,000 test images. Several representative baselines are selected as Network in Network (NIN) \cite{Lin2013Network}, FitNets \cite{Rao2016Noise}, Deep Supervised Network (DSN) \cite{Lee2014Deeply}, High-Way \cite{Srivastava2015Training}, All-CNN \cite{Springenberg2014Striving}, Exponential Linear Units (ELU) \cite{DjorkArn2015Fast}, FitResNets \cite{Mishkin2015All}, gcForest \cite{Zhou2017Deep} and Deep ResNet \cite{He2016Deep}. We could conclude that:
\begin{table}[t]
	\caption{Performance Evaluation (Error (\%)) on CIFAR.}
	\centering
	\label{tab:paraphrase}
	\renewcommand\arraystretch{1.1}
	\begin{tabular}{c c c}
		\Xhline {1.2pt} Methods & CIFAR-10 & CIFAR-100 \\
		\Xhline {1.2pt}
		NIN & 8.81 & 35.68 \\
		DSN & 8.22 & 34.57 \\
		FitNets & 8.39 & 35.04 \\
		High-Way & 7.72 & 32.39 \\
		All-CNN & 7.25 & 33.71 \\
		ELU & 6.55 & 24.28 \\
		FitResNets & 5.84 & 27.66 \\
		ResNet & 6.61 & 25.16 \\
		gcForest & 31.00 & - \\
		Random Forest & 50.17 & - \\
		\hline
		Single Target Network & - & 89.37 \\
		\hline
		\hline 
		\textbf{NDT (depth = 4)} & - & \textbf{84.52} \\
		\Xhline {1.2pt}
	\end{tabular}
\end{table}

\begin{figure*}
	\centering
	\includegraphics[width=0.9\linewidth]{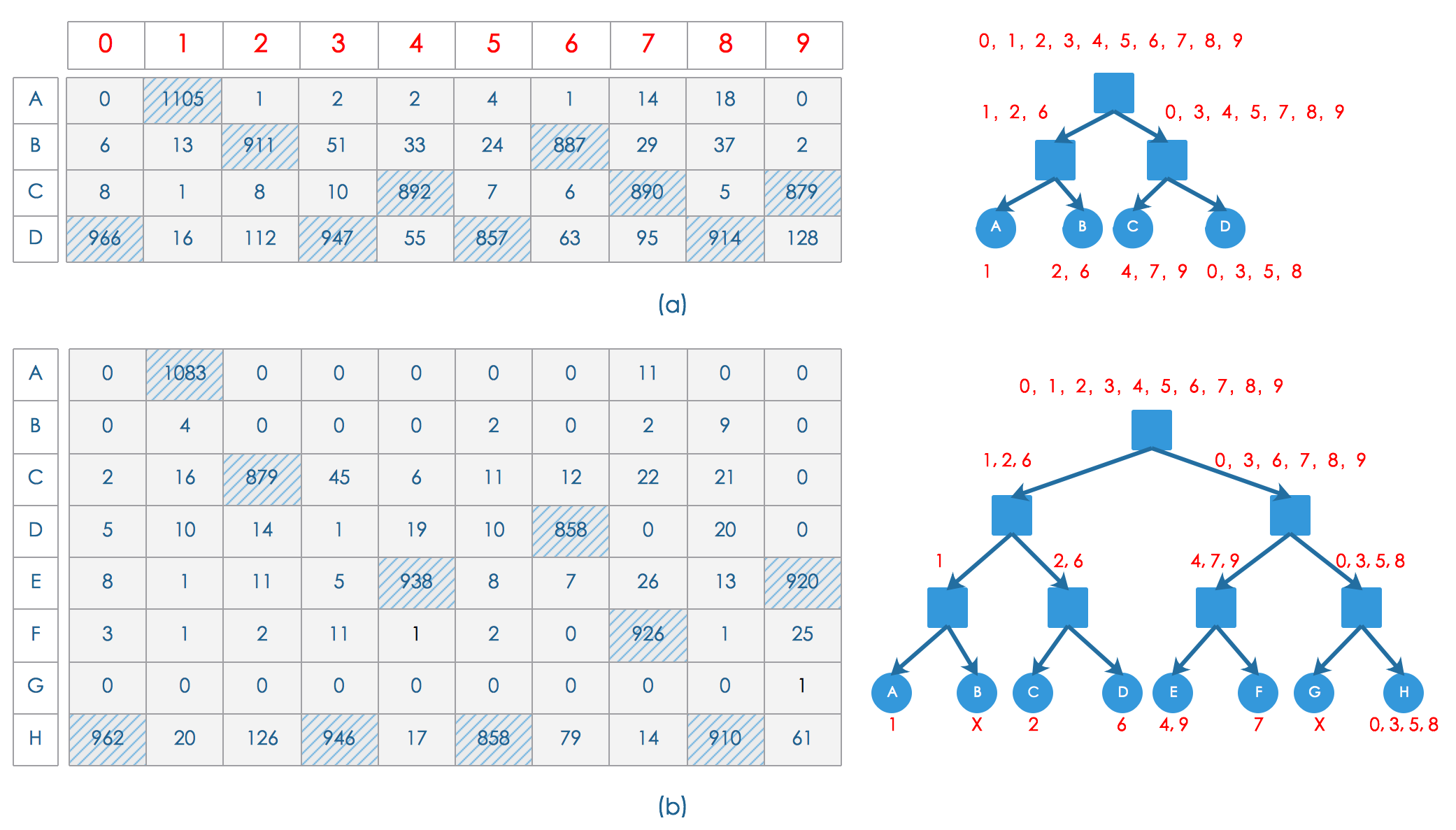}
	\caption{Case Study for NDT in MNIST with depth = 2 (a) and depth = 3 (b). The left tables are the test sample numbers that correspond to $row$-th leaf node and $col$-th category. For example, the sliced ``1105'' means there are 1105 test samples of category ``1'' in leaf node ``A''. We slice the main component of a leaf and draw the corresponding decision trees in the right panel. Notably, ``X'' indicates the empty class.}
	\label{fig:casestudy}
\end{figure*}

\begin{enumerate}
	\item NDT will beat all the strong baselines, which verifies the effectiveness of neural decision trees and justifies the theoretical analysis.
	\item  Compared to single target network, NDT promotes 4.85 point, which illustrates the ensemble of target network is effective.
	\item Compared with gcForest, the performance improves - points, which illustrates that neurons empower the decision trees more effectively than direct ensembles.
	\item Compared with ResNet that is the strongest baseline, we promote the results over - points, which justifies our assumption, that NDT could reduce the category number of leaf nodes to enhance the intelligence systems.
\end{enumerate}

\subsection{Case Study}
To further testify our assumption that NDT could reduce the category number of leaf nodes, we perform a case study in MNIST. We make a statistics of test samples for each leaf node, illustrated in Figure \ref{fig:casestudy}. The item of table means $row$-th leaf node has how many samples in $col$-th category. For example, the ``1105'' in the first row and second column, means that there are 1,105 test samples of category ``1'' are pre-classified into leaf node ``A''. Correspondingly, we draw the decision trees in the right panel with labeled categories, which specifically illustrates the decision process of NDT. For a complete verification, we vary the depth of NDT with 2 and 3. 

Firstly, we could clearly draw the conclusion from Figure \ref{fig:casestudy}, that each leaf node needs to predict less categories, which justifies our assumption. For example,  in the bottom figure, the node ``A'' only needs to predict the category ``1'', which is a single classification, and the node ``H''  only needs to predict the categories ``0,3,5,8'' which is a four classification. Because small classification is less difficult than large one, our target network in the leaf could perform better, which leads to performance promotion in a tree-ensemble manner.

Secondly, from Figure \ref{fig:casestudy}, split purity could be worked out. Generally, the two-layer tanh multi-perception achieves a decent split purity. Indeed, the most difficult leaf nodes (e.g. \textit{``D'' in the top and ``H'' in the bottom}) are not perfect, but others gain a competitive split purity. Statistically, the main component or the sliced grid takes 92.4\% share of total samples, which in a large probability, NDT would perform better than 92.4\% accuracy in this case.

Finally, we discuss the hyper-parameter $depth$. From the top to the bottom of Figure \ref{fig:casestudy}, the categories are further split. For example, the node ``B'' in the top is split into ``C'' and ``D'' in the bottom, which means that the category ``2'' and ``6'' are further pre-classified. In this way, deep neural decision tree is advantageous. But much deeper NDT makes less sense, because the categories have been already split well. There would be mostly no difference for 1- or 2-classification. However, considering the efficiency and consuming resources, we suggest to apply a suitable depth, or theoretically about $log_2(|C|)$, where $|C|$ is the total category number.


\section{Related Work}
In this section, we briefly introduce three lines of related work: image recognition, decision tree and neural graph.

Convolution layer is necessary in current neural architectures for image recognition. Almost every model is a convolutional model with different configurations and layers, such as  All-CNN \cite{Springenberg2014Striving} and DSN \cite{Lee2014Deeply}. Empirically, deeper network produces better accuracy. But it is difficult to train much deeper network for the issue of vanishing/exploding gradients, \cite{glorot2010understanding}. Recently, there emerge two ways to tackle this problem: High-Way \cite{Srivastava2015Training} and Residual Network \cite{He2015Deep}. Inspired by LSTM, high-way network applies transform- and carry-gates for each layer, which allow information to flow across layers along the computation path without attenuation. For a more direct manner, residual network simply employs identity mappings to connect relatively top and bottom layers, which propagates the gradients more effectively to the whole network. Notably, achieving the state-of-the-art performance, residual network (ResNet) is the strongest model for image recognition, temporarily.

Decision tree is a classic paradigm of artificial intelligence, while random forest is the representative methodology of this branch. During recent years, completely random tree forest has been proposed, such as iForest \cite{Liu2008Isolation} for anomaly detection. However, with the popularity of deep neural network, lots of researches focus on the fusion between neurons and random forest. For example, \cite{Richmond2015Relating} converts cascaded random forests to convolutional neural network, \cite{welbl2014casting} leverages random forests to initialize neural network. Specially, as the state-of-the-art model, gcForest \cite{Zhou2017Deep} allocates a very deep architecture for forests, which is experimentally verified on several tasks. Notably, all of this branch could not jointly train the neurons and decision trees, which is the main disadvantage.
 
To jointly fuse neurons and logics, \cite{xiao2017hungarian} proposes the basic principle of neural graph, which could embed traditional logics-based algorithms into neural architectures. The seminal paper merges the Hungarian algorithm with neurons as Hungarian layer, which could effectively recognize matched/unmatched sentence pairs. However, as a special case, the variables only introduced in the condition could not be updated, which is a disadvantage for characterizing complex systems. Thus, this paper focuses on this issue to make a fully functioned neural graph.

\section{Future Work}
We list three lines of future work: design new components of neural graph, implement a script language for neural graph and analyze the theoretical properties of learnable Turing machine.

This paper exemplifies an approach to embed decision tree into neural architectures. Actually, many traditional algorithms could promote intelligence system with neurons. For example, neural A$^*$ searching could learn the heuristic rules from data, which could be more effective and less resource consuming. For a further example, we could represent the data with deep neural networks, and conduct label propagation upon the hidden representations, where the propagation graph is constructed by K-NN method. Because the label propagation, K-NN and deep neural networks are trained jointly, the performance promotion could be expected.

In fact, a fully functioned neural graph may be extremely hard and complex to implement. Thus, we expect to publish a script language for modeling neural graph and also a library that includes all the mainstream intelligence methods. Based on these instruments, neural graph could be more convenient for practical usage.

Finally, as we discussed, neural graph is Turing complete, making a learnable Turing machine. We believe theoretical analysis is necessary for compilation and ability of neural graph. Take an example. Do the learnable and static Turing machine have the same ability? Take a further example. Could our brain excel Turing machine? If not, some excellent neural graphs may gain advantages over biological brain, because both of them are learnable Turing machines. If it could, the theoretical foundations of intelligence should be reformed. Take the final example. What is the best computation model for intelligence?

\section{Conclusion}
This paper proposes the principle of fully functioned neural graph. Based on this principle, we design the neural decision tree \textbf{(NDT)} for image recognition. Experimental results on benchmark datasets demonstrate the effectiveness of our proposed method.

\newpage

\bibliography{Ref}
\bibliographystyle{icml2018}

\end{document}